\documentclass{article}
\usepackage{ijcai24}

\usepackage{times}
\usepackage{soul}
\usepackage{url}
\usepackage[hidelinks]{hyperref}
\usepackage[utf8]{inputenc}
\usepackage[small]{caption}
\usepackage{graphicx}
\usepackage{amsmath}
\usepackage{amsthm}
\usepackage{booktabs}
\usepackage{algorithm}
\usepackage{algorithmic}
\usepackage[switch]{lineno}

\usepackage[symbol]{footmisc}
\title{Interpretable Network Visualizations: A Human-in-the-Loop Approach for Post-hoc Explainability of CNN-based Image Classification}

\author{
Matteo Bianchi\footnotemark[1]\and
Antonio De Santis\footnotemark[1]\and
Andrea Tocchetti\And
Marco Brambilla\\
\affiliations
Politecnico di Milano, DEIB, I-20133 Milano, Italy\\
\emails
\{firstname.lastname\}@polimi.it
}

\begin{document}

\maketitle

\footnotetext[1]{Equal contribution}
\renewcommand*{\thefootnote}{\arabic{footnote}}

\begin{abstract}
    Transparency and explainability in image classification are essential for establishing trust in machine learning models and detecting biases and errors. State-of-the-art explainability methods generate saliency maps to show where a specific class is identified, without providing a detailed explanation of the model's decision process. Striving to address such a need, we introduce a post-hoc method that explains the entire feature extraction process of a Convolutional Neural Network. These explanations include a layer-wise representation of the features the model extracts from the input. Such features are represented as saliency maps generated by clustering and merging similar feature maps, to which we associate a weight derived by generalizing Grad-CAM for the proposed methodology. To further enhance these explanations, we include a set of textual labels collected through a gamified crowdsourcing activity and processed using NLP techniques and Sentence-BERT. Finally, we show an approach to generate global explanations by aggregating labels across multiple images.
\end{abstract}

\section{Introduction}
In recent years, Deep Neural Networks (DNNs) have transformed the field of Artificial Intelligence by revolutionizing the way machines learn. Convolutional Neural Networks (CNNs) have emerged as the state-of-the-art for image classification tasks \cite{Krizhevsky}, thanks to their ability to recognize patterns and features. However, as AI models have become more powerful, their decision-making process has become increasingly complex and less transparent. As a result, the use of the term \textit{black-box} has become prevalent to describe such models since only their input and output are known, while their internal workings are too intricate for humans to comprehend. This issue leads to opacity in AI decisions, a serious concern \cite{Eschenbach} that causes a loss of trust towards these systems \cite{Lipton}. Furthermore, debugging black-box models is challenging without insights into how a model generates its predictions.
The increasing need for transparent AI led to the rise of a research area called Explainable Artificial Intelligence (XAI). Researchers made significant progress in developing techniques for producing explanations of AI decisions, although some limitations still need to be addressed. For image classification, these techniques mainly focus on producing saliency maps that highlight the regions of the input image contributing the most to the output. While this suggests whether the AI is looking at the ``right thing'', it doesn't explain which features guided the model into predicting the correct class.

In this article, we present an explainability framework that combines XAI with human knowledge to create both \emph{local} and \emph{global explanations} for any CNN model, without requiring any modification or performance trade-off. The main objective of our method is to offer a detailed view of the features the CNN extracts from the input image and their respective importance. We apply a layer-wise clustering algorithm to the feature maps to identify which neurons are focusing on the same region of the input image.
These features are then represented as saliency maps, computed by merging similar feature maps, and a descriptive label collected through a gamified crowdsourcing application to further enhance interpretability \cite{Szymanski}.
The result is a layer-wise visualization of the CNN's feature extraction process for a single input image, a local explanation we refer to as Interpretable Network Visualization (INV).
Furthermore, textual labels facilitate the aggregation of multiple features across different images, which can be used for generating class-level global explanations that provide insights into how a model recognizes a particular class.

\section{Background and Related Works}
While there is no universal agreement on the definition of XAI, it can be defined as an \emph{``interface between humans and a decision-maker that is, at the same time, both an accurate proxy of the decision-maker and comprehensible to humans''} \cite{Arrieta}. In this article, we focus on post-hoc explainability, which aims at explaining the outputs of black-box models without modification or re-training \cite{Xu}.
The scientific literature classifies post-hoc explainability into two main categories: model-agnostic and model-specific \cite{Guidotti}. Model-agnostic approaches refer to XAI methods that are independent of the underlying ML model. These methods don't access the internal workings of a model, but instead focus on studying the input-output relationship. One of the most notable contributions to the model-agnostic approach is Local Interpretable Model-Agnostic Explanations (LIME) \cite{lime}, which produces visual explanations by perturbing the input to compute the importance of different regions of the image. A similar approach is SHapley Additive exPlanations (SHAP) \cite{shap}, which uses shapley values to compute the importance of each pixel of an input image.

On the other hand, model-specific approaches apply to specific categories of ML models. Concerning CNN architectures, Simonyan \emph{et al.} \shortcite{Simonyan} were among the first to generate visual explanations based on the gradients of the output w.r.t. each pixel, whose importance is represented in a saliency map.
A different approach, proposed by Zhou \emph{et al.} \shortcite{Zhou}, extracts feature maps to generate saliency maps by substituting the fully connected layers with a Global Average Pooling (GAP) layer. The weights obtained by re-training the GAP layer represent the importance of each feature maps towards a specific class. Finally, a Class Activation Map (CAM) is obtained as the weighted sum of the feature maps. Later, Selvaraju \emph{et al.} \shortcite{Selvaraju} introduced a generalized version of CAM called Gradient-weighted Class Activation Mapping (Grad-CAM), which can be applied without modifying the network. The main idea behind Grad-CAM is that the weights can be computed by using a GAP operation directly on the gradients of the output for a given class w.r.t. the feature maps of the last convolutional layer.
However, according to Sundararajan \emph{et al.} \shortcite{Sundararajan_imp_inv}, relying solely on gradients can sometimes result in an inaccurate assessment of feature importance due to the saturation of gradients. To address this, Sundararajan \emph{et al.} \shortcite{Sundararajan_imp_inv} introduced Integrated Gradients (IG), a method that computes the contribution of each pixel by integrating the gradients along a path from a baseline (i.e., a black image) to the actual input image. Similarly to SHAP, these contributions are used to produce fine-grained saliency maps.

The main limitation of saliency methods is that their interpretation is subjective. For instance, suppose Grad-CAM highlights the region of a cat’s nose as important for the prediction ``cat''. This might suggest that the network uses information about the nose to make the prediction, however, it could be using information about the texture of the fur or the presence of whiskers to make its prediction. 
A solution was proposed by Kim \emph{et al.} \shortcite{Kim} with a framework named Testing with Concept Activation Vectors (TCAV). This method computes the importance of a user-defined feature for predicting a class by analyzing the network's activations produced by a set of sample images representing that feature. For instance, one can use images of stripes to determine whether the network is using the stripes feature to recognize the zebra. This method can detect specific biases in neural networks (e.g., ethnicity-related) and can be considered complementary to saliency methods as it provides a global understanding of the model behavior, while saliency methods provide explanations for specific predictions. Both approaches can be very effective in gaining insights into AI decisions, although it has been shown they can also introduce bias in the explanations. For saliency methods, this happens when humans misinterpret why a region is highlighted due to biased assumptions. For TCAV, the bias can be introduced by the images selected for representing a feature \cite{Tong}.

\begin{figure*}
    \centering
    \includegraphics[width=1\textwidth]{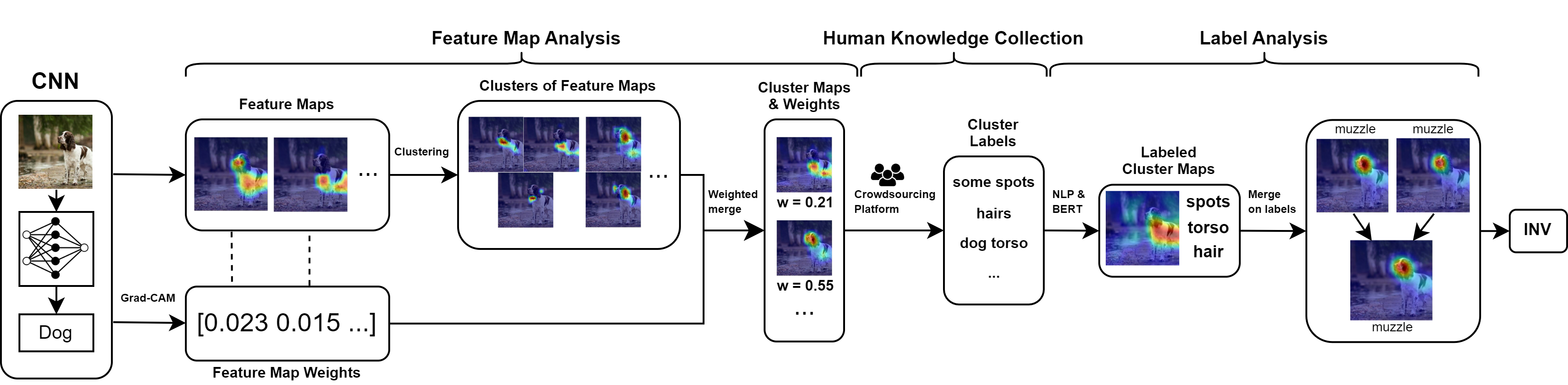}
    \caption{A pipeline showing the process of generating an INV. In the first step, feature maps and their weights are extracted from the CNN. These feature maps are clustered to generate cluster maps. Subsequently, labels are collected through crowdsourcing and processed using NLP techniques. Finally, cluster maps with the same top label are merged.}
    \label{fig:pipeline}
\end{figure*}

\subsection{Human Knowledge and Explainable AI \label{crowdsourcing}}
Despite the advancements of XAI, there are still limitations in ensuring that the explanations are understandable to humans. For this reason, researchers \cite{tocchettirole,WangCrowdsouricng,Magister,Castro} turned to human-in-the-loop techniques that utilize human reasoning to improve explanations of AI models. For instance, Mishra and Rzeszotarski \shortcite{Mishra} developed a crowdsourcing method for collecting concepts for image classification explanations by asking users to outline the location of the features needed to identify a class. Later, Tocchetti \emph{et al.} \shortcite{Tocchetti} proposed a two-player gamified crowdsourcing activity for collecting human concepts that can be used for explainability purposes. In the proposed activity, one player is asked to guess the entity in a picture without seeing it by asking about its features through closed questions to the other player, who provides the answers and takes note of the guessed features while outlining them on the image.

Among the researchers who employed human knowledge in conjunction with XAI methods, Lu \emph{et al.} \shortcite{Lu} proposed a game based on ``Peek-a-Boom'' to evaluate explanations generated by different XAI techniques. In their implementation, only a small part of an image is initially shown, starting from the region deemed the most important by a saliency method. If the player cannot guess the picture, more pixels are revealed. The number of pixels the player needs to guess correctly represents how well humans can interpret the explanation. Another approach that, instead, focuses on global explainability was introduced by Balayn \emph{et al.} \shortcite{Balayn}. They suggested augmenting saliency maps by incorporating semantic concepts through crowdsourcing annotations. The main advantage is that annotations can be aggregated to generate global explanations. Overall, these methods demonstrate the value of incorporating human knowledge in the explainability of ML models, thus foreseeing a promising direction for the field of XAI.

\section{Interpretable Network Visualizations}
Achieving a comprehensive explanation of an AI decision requires examining the entire AI decision process. In the case of CNNs for image classification, this process includes multiple feature extraction stages throughout multiple layers. For example, considering a dog's image, one layer might identify the shape, the subsequent could extract more specific features such as ears or eyes, and the final layer could identify the dog's head and body. Based on this idea, we propose an explainability framework, named Interpretable Network Visualization (INV), to generate local explanations in the form of a layer-wise overview of the features extracted by the network. Each column of an INV represents a layer and consists of a set of saliency maps providing a visual representation of the areas of the input image where important features were identified. Each saliency map represents a group of feature maps clustered by similarity (i.e., focusing on the same image region). The visualization also includes the relative importance of each feature w.r.t. the predicted class. In addition, saliency maps are linked to a set of crowdsourced labels that describe the highlighted features and are intrinsically understandable since they are provided directly by humans. We use a set of labels instead of a single one, as feature maps may focus on a combination of features. For instance, a saliency map could highlight a person and the sky, as this combination leads the network to predict the ``parachute'' class rather than any of these two features individually.

An INV can be built considering all layers of the CNN or a selected subgroup of interest. The most efficient approach is to focus on deeper layers as they contain more semantic concepts since their receptive fields are bigger and they are closer to the output. In contrast, shallow layers focus on detecting less discriminative features such as basic shape information (e.g., edges and outlines) \cite{Zeiler}.
Given a CNN trained for image classification and an input image, these are the steps to build an INV, as outlined in Figure \ref{fig:pipeline}:
\begin{enumerate}
  \item \textit{Feature Maps Analysis}: Feature maps are extracted and clustered. Clusters are merged to generate the saliency maps representing the features extracted by the CNN.
  \item \textit{Human Knowledge Collection}: Labels are collected through crowdsourcing to associate textual descriptions with the extracted features.
  \item \textit{Label Analysis}: Collected labels are processed using Sentence-BERT and other NLP techniques to identify synonyms and remove errors.
\end{enumerate}

\begin{figure*}
    \centering
    \includegraphics[width=0.781\textwidth]{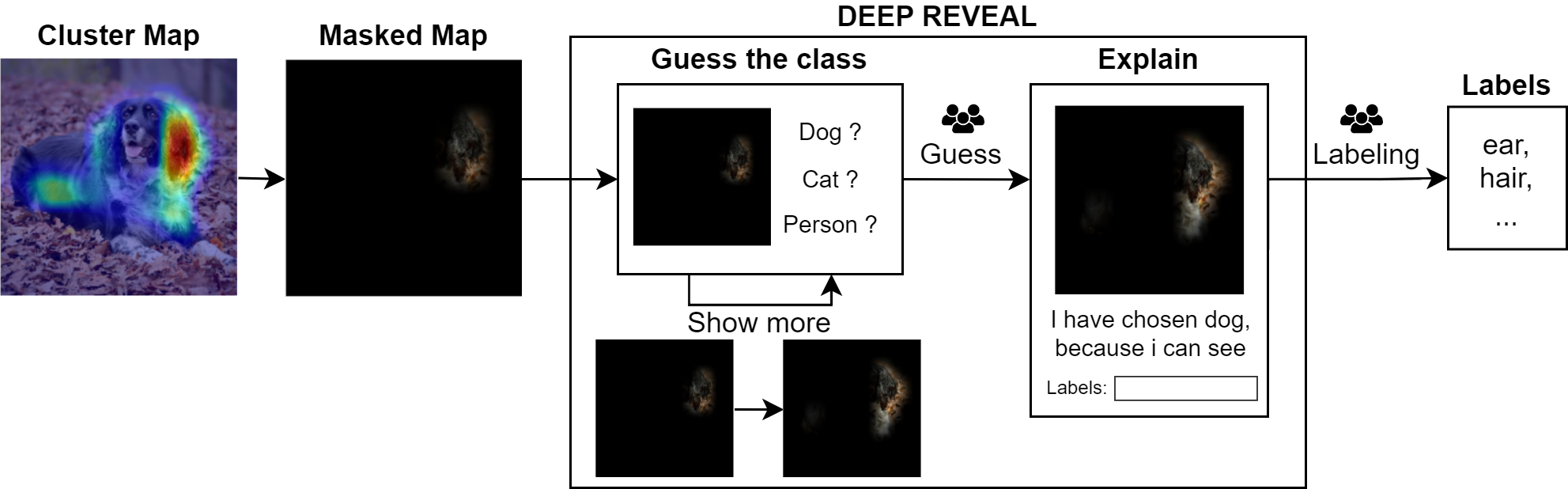}
    \caption{A pipeline describing the label collection process through \textit{Deep Reveal}. The masked version of the cluster map is shown to users who can try to guess right away or increase the visible portion of the image. After guessing, users provide labels to explain their decision.}
    \label{fig:crowd_pipeline}
\end{figure*}

\subsection{Feature Maps Analysis \label{method:fmap_and_weights}}
Feature maps are extracted and associated with their class-specific weights towards the predicted class. We use the Grad-CAM algorithm to compute the weight since it is a straightforward approach applicable to all CNN architectures.
We then apply a layer-wise unit normalization to the weights so that we can interpret them as a measure of relative importance. Feature maps with weights equal to or less than zero are removed. A positive weight threshold can optionally be selected to exclude feature maps with low importance, which can further optimize the clustering of important features. Such threshold depends on the number of feature maps, hence it may vary based on the network and the layer of interest. The idea behind thresholding is to remove noisy feature maps while retaining the vast majority (70-90\%) of the total weight. 
Before clustering, we perform min-max normalization and dimensionality reduction. For the latter, we use a combination of two techniques: Principal Component Analysis (PCA) \cite{Maćkiewicz} and t-distributed Stochastic Neighbor Embedding (t-SNE) \cite{Maaten}. More specifically, we apply PCA to reduce the number of dimensions to a reasonable amount (30-50) before using t-SNE. 
Subsequently, we apply Agglomerative Hierarchical Clustering for each layer of the network, using the Euclidean distance and ward-linkage. This clustering algorithm is more suitable for our problem than density-based and centroid-based approaches as the former removes noisy points that could be important in our analysis while the latter performs better under specific assumptions (e.g., spherical distribution of variables) that are unlikely to be verified. Additionally, hierarchical clustering allows choosing the number of clusters. This is important because this value should not be overwhelmingly high as the ultimate objective is to produce human-understandable explanations. Generating too many clusters can also be problematic during the labeling phase as it would require a substantial number of participants. For these reasons, a reasonable number of clusters (e.g., from 3 to 8) must be set based on model size, availability of crowdsourcing resources, and human comprehensibility.
Furthermore, the optimal number of clusters can differ for each layer as it depends on the number of features a convolution layer extracts for a specific input image, which can't be known beforehand. Therefore, we used the average silhouette score (i.e., a measure of cohesion and separation of clusters) to select the optimal number of clusters for a given input image and layer.
Once the clustering process is complete, each cluster's feature maps are merged using a weighted average. This produces saliency maps, which we refer to as cluster maps, representing an entire cluster of feature maps. Each cluster map is assigned a weight that defines its importance towards the predicted class. This value is computed by summing the weights of all feature maps belonging to that cluster. More formally, these weights are derived from the Grad-CAM equation (\ref{cam_merging_1}), which applies to any layer. Considering $n$ clusters, by grouping the terms $w_jA^j$ that correspond to feature maps belonging to a specific cluster $C_i$, we get,
\begin{equation}
    L_{Grad-CAM}=\sum_kw_kA^k=\sum_{i=1}^{n}\sum_{j\in C_i}w_jA^j
    \label{cam_merging_1}
\end{equation}
From (\ref{cam_merging_1}), we can factor out the sum of the weights to obtain,
\begin{equation}
\sum_{i=1}^{n}\Biggl(\,\underbrace{\biggl(\sum_{j\in C_i}w_j\biggr)}_{w_{C_i}}\cdot\overbrace{\frac{\sum_{j\in C_i}w_jA^j}{\sum_{j\in C_i}w_j}}^{A_{C_i}}\,\Biggr)=\sum_{i=1}^{n}w_{C_i}A_{C_i}
\label{cam_merging_2}
\end{equation}
The term $w_{C_i}$ represents the cluster weight, multiplied by a term $A_{C_i}$ that is the cluster map obtained through the weighted average. When using this method to compute the weights, the Grad-CAM map can still be derived by merging cluster maps. Hence, they can be seen as condensed feature maps. After obtaining a set of cluster maps for every layer and their corresponding weight, we can optionally choose a threshold to exclude cluster maps with low overall weight. 

\subsection{Human Knowledge Collection \label{methodology_humans}}
This step aims to collect labels describing the features highlighted in each cluster map. To ensure these labels are human-understandable and unbiased, they are obtained through a crowdsourcing activity. This enables associating visual explanations with concepts that are familiar to and interpretable by humans. Before describing the crowdsourcing activity, we must address the content to be provided to participants when labeling cluster maps. The general approach to obtain an interpretable visualization of a feature map is to generate an overlay between the input image and the feature map after up-scaling, blurring, and color mapping. However, directly labeling these overlays has a problem: if humans know the image's subject, they will most likely lose focus on the highlighted areas. Hence, the labels might not accurately describe the highlighted features. For example, a cluster map highlighting a portion of grass on a soccer field may be labeled as ``soccer field'' instead of ``grass''. Although the label ``soccer field'' is acceptable, ``grass'' is more detailed and describes only the highlighted area. Hence, we hide non-highlighted portions to prevent such behaviors. This is achieved by computing a mask (i.e., a binary image) that defines which pixels to show. Then, a masked image is obtained by overlaying a blurred mask on top of the input image (refer to Figure \ref{fig:crowd_pipeline}).

For collecting the labels, we employ a gamified crowdsourcing activity since we want participants to observe and analyze features to guess the correct class. similarly to what a neural network does. Additionally, gamification can significantly increase engagement, leading participants to put more effort into their responses, resulting in higher-quality labels \cite{Morschheuser}. The actual activity consists of a custom web-based game called \textit{Deep Reveal} (whose process is reported in Figure \ref{fig:crowd_pipeline}) in which users are presented with the masked image of a cluster map for which they are required to guess the class from a subset of the model's classes chosen at random, and then explain the discerning factors that led to their decision.
We avoid deterministic approaches to select the subset, such as considering the prediction confidences, as this may introduce bias and lead to recognizable patterns for users.
Similarly to Peek-a-Boom \cite{Lu}, users can obtain hints by gradually increasing the displayed area. Such a mechanism is implemented by using multiple masked images with a gradually increasing percentile. Users can increase the displayed area up to five times (i.e., generating five additional masks), after which the game gives them the option to resign.
Once users select an option, the game prompts them to specify which features helped them guess the class. These inputs are then used as labels for the cluster maps. We ask participants to guess the image before labeling to focus their attention on the most discriminative features. To avoid introducing biases, players should not be provided with a list of labels to choose from. This prevents users from inferring something they would not have otherwise seen if provided with multiple options. Furthermore, the game includes a scoring system and a leaderboard to increase engagement and competition. In particular, the more users reveal the image, the fewer points they earn for each game. Additionally, the game includes a set of screening masked images, allowing to filter out data from untrustworthy users.

\subsection{Label Analysis \label{data_analysis}}
The collected data may contain errors and impurities (e.g., long phrases, synonyms, stop-words, misspelled words) since users are allowed to insert multi-word labels freely. 
To address these issues, we initially split the labels based on the guessed class and remove the class name from the labels, as it does not provide any valuable information and could potentially lead to information loss. For instance, an NLP technique might consider the labels ``ear of the dog'' and ``dog paws'' as similar sentences despite referring to different features.
Subsequently, we convert these labels into word embeddings using \emph{all-mpnet-base-v2}, a Sentence-BERT model \cite{bert}. Following this, we perform Agglomerative Clustering using cosine similarity, complete-linkage, and the average silhouette score to select the optimal number of clusters. Then, we use cosine similarity to map these clusters to a single-word label by identifying the word that is the nearest to the cluster’s embeddings, from all words within the cluster (excluding stop-words). Finally, groups represented by labels with the same lemma are unified.

The next step is to heuristically assign a score to each label to evaluate which ones most accurately describe the cluster map. This score is calculated as the label frequency minus a penalty of 0.1 for each hint used by players who assigned that label. When a label is provided only by users who guessed the image wrongly, its score counts only as one-fourth since they have a higher probability of being imprecise. Furthermore, two or more cluster maps may represent the same feature and could be labeled similarly. This can happen due to imperfections in the feature maps' clustering process or because the same feature is present in different image regions. To address this, we perform a layer-wise merge of clusters with the same top labels, as they focus on the same feature. The cluster maps are merged through a weighted average and by summing their weights. The labels' scores are also combined through a weighted average. Finally, the INV of an image can be generated by organizing its cluster maps together with their weights and labels into a layer-wise visualization.

\begin{figure*}
    \centering
    \includegraphics[width=1\textwidth]{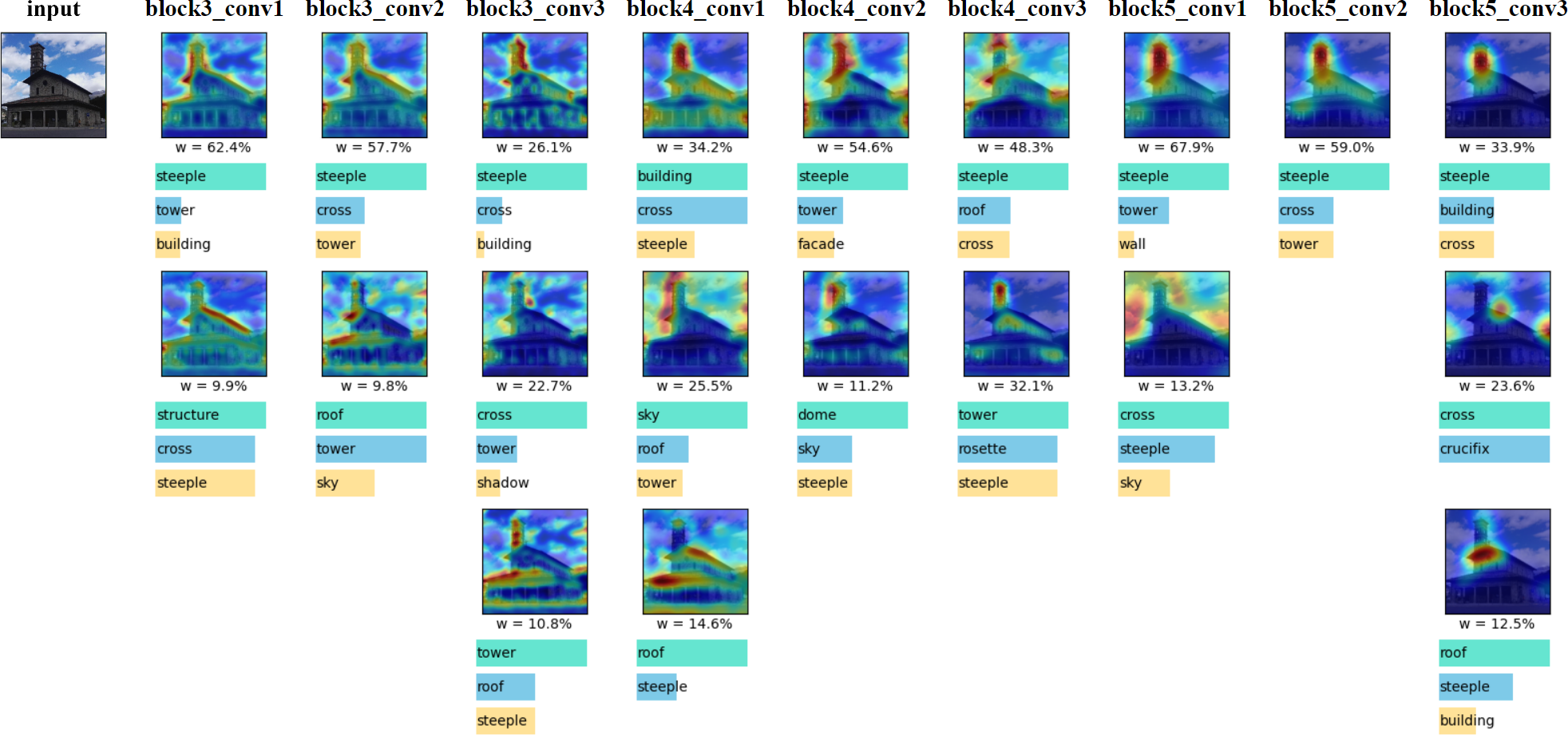}
    \caption{An INV for an image of the class ``church''. The most important feature for this prediction was ``steeple''. However, it can be observed that other elements, such as ``cross'' and ``roof'' also contributed to the classification.}
    \label{fig:church_local}
\end{figure*}

\section{Experiment Setup}
We designed an image classification model based on the standard VGG-16 architecture initialized with the ImageNet weights. We trained the model using \emph{Imagenette}\footnote{\url{https://github.com/fastai/imagenette}} dataset, a small subset of ImageNet consisting of ten classes.
In our experiment, we considered a set of 50 predictions, consisting of five correctly classified images per class, and extracted the feature maps for the last nine convolutional layers, as shallower ones primarily focus on less discriminative features such as shapes, edges, and outlines. 
We applied two thresholding steps, as described in previous sections. The first occurs before clustering, removing feature maps with a low weight. The second one is applied after clustering, removing clusters with low total weight using a heuristic formula (\ref{c_threshold}). 
\begin{equation}
    th=max(max(weights)/3,\ avg(weights)/2)
    \label{c_threshold}
\end{equation}
The results of thresholding and clustering are presented in Table \ref{tab:clustering_res}.
The proportion of removed feature maps was significantly higher in deeper layers, while the total weight left over was relatively constant. This suggests that the weights are more condensed in deeper layers, allowing the aggregation of more relevant information in fewer clusters. Finally, we merged the feature maps of each cluster, resulting in a total of 1954 cluster maps to be labeled.
For the labeling phase, we implemented \textit{Deep Reveal} and shared it with 210 participants, the majority of which were computer engineering students with various cultural backgrounds. 
We allowed them to insert labels both in Italian and English. For this reason, we performed a translation step before processing the labels, which was done using \emph{DeepL API}, with the guessed class as a translation context.
\begin{table}
    \centering
    \fontsize{9.9pt}{9.9pt}\selectfont
    \begin{tabular}{lrrrr}
    %{}              & {}     & average & average & average\\
    {}          & fmaps     & leftover & leftover & leftover  \\
    layer name & threshold & clusters & fmaps & weight \\
    \midrule
    \rule{0pt}{10pt}block3\_conv1     &              $<$0.35\% &   4.3  &  52.7\% &        70.2\% \\           
    block3\_conv2     &                $<$0.35\%    &   4.4 &  51.4\% &         68.6\% \\         
    block3\_conv3     &                $<$0.35\%  &    4.8  &  46.2\% &          65.2\% \\            
    block4\_conv1     &               $<$0.175\% &   4.2  & 46.5\% &          71.9\% \\             
    block4\_conv2     &                $<$0.175\%   &    4.8 & 44.9\% &          72.1\% \\            
    block4\_conv3    &               $<$0.175\%    &   5.2& 35.9\% &          72.6\% \\            
    block5\_conv1    &                $<$0.175\%  &   4.6 &  30.4\% &          73.8\% \\
    block5\_conv2    &                $<$0.175\%  &   3.6 & 25.4\% &          72.5\% \\            
    block5\_conv3    &               $<$0.175\%   &   3.3 & 14.6\% &          78.4\% \\             
    \end{tabular}
    \caption{The table presents the results of thresholding and clustering, with the threshold being lower in deeper layers due to the presence of twice as many feature maps. The values for leftover clusters, feature maps (fmaps), and weight are averaged across all images.}
    \label{tab:clustering_res}
\end{table}
\section{Results and Discussion}
We conclude by presenting the results of applying the INV framework. We discuss the INVs obtained from the experiment and compare them with state-of-the-art XAI techniques. Subsequently, we propose a method to aggregate features extracted through INVs across various images to produce class-level global explanations.
All results for both the INVs and global explanations are available in our GitHub repository\footnote{\url{https://github.com/Antonio-Dee/interpretable-network-visualizations}}. Finally, we discuss \emph{Deep Reveal}'s usability and workload.

\subsection{INVs Evaluation\label{abstract_networks_results}}
Following the INV structure, we organized the cluster maps into layers, displaying their weight and top three labels. Figure \ref{fig:church_local} shows an INV for an image of class ``church''. In this case, the prediction was determined by the features ``steeple'', ``cross'', and ``roof'', with the former being the most important.
As the layers get deeper, the focus on these features increases, which is an expected behavior since they are the most discriminative ones. 
Furthermore, it can be observed that shallower layers interpret the same features differently than deeper ones, placing more emphasis on their shape and outlines. This is particularly noticeable for ``roof'' and ``steeple'' in the provided example.

Although INVs provide detailed visualizations of the features extracted by the network, they also have limitations. In particular, labels may not always describe the cluster map precisely since they are provided by humans who are trying to guess the class, especially for less discriminative features (e.g., in Figure \ref{fig:church_local}, the second cluster map of layer \emph{block5\_conv1} is focusing on ``sky'', more than ``steeple'' or ``cross'').
Another limitation is derived from the application of NLP techniques to automate the process of Label Analysis. This is because humans can use words with different meanings to refer to the same visual entity (e.g., ``logo'' and ``brand'') and even robust models like Sentence-BERT might not account for such instances. For example, when examining Figure \ref{fig:church_local}, one can notice that in the final layer of the INV, the labels ``cross'' and ``crucifix'' are distinct, despite being more reasonable to combine them into a single label since they represent the same visual entity. A similar situation is observed with ``tower'' and ``steeple'' in earlier layers.

\subsubsection{Comparative Analysis with Human Subjects}
We conducted a survey aimed at comparing INVs with state-of-the-art local XAI methods to objectively assess the results derived from our experiment. For this comparison, we chose Grad-CAM, LIME, and SHAP. Specifically, we used SHAP’s Gradient Explainer implementation, which uses Expected Gradients \cite{EIG}, an extension of Integrated Gradients, to estimate Shapley values. We also incorporated a simplified version of INV that only includes the final layer and a single label per cluster map.
For each XAI method, we provided participants with a brief description of the technique, along with four instances of explanations generated on our CNN, each from a different class. We utilized the same images for all methods, which were randomly selected from the pool of 50 images used in the experiment. Regarding the survey questions, we followed the approach proposed by Aechtner \emph{et al.} \shortcite{comparison}. We evaluated the following five aspects on a 7-point Likert scale: \emph{understandability}, \emph{usefulness}, \emph{trustworthiness}, \emph{informativeness}, and \emph{satisfaction}. We also asked the participants to specify whether they have a background in AI or not.
The questionnaire was completed by 165 participants, of which 71\% had a background in AI.
%Nonetheless, the responses from the two groups did not show any significant differences.
Concerning the results shown in Table \ref{tab:questionnaire_xai}, a trend is noticeable where SHAP and LIME consistently underperformed compared to INVs and Grad-CAM. Although our methods performed well in all aspects, a t-test revealed a significant difference (p-value $<$ 0.05) only in the scores of INV against all other methods in terms of \emph{informativeness}. The simplified version of INV also outperformed state-of-the-art methods significantly. However, for other aspects, there was no statistical difference between INV, simplified INV, and Grad-CAM. This implies that our methods are superior in \emph{informativeness} while being at least equal in other aspects.

\begin{table*}
    \centering
    \begin{tabular}{lcccccc}
    %& {} & {} & {} & (AI Novices) & (AI Experts) & {} \\
     & Understandability & Usefulness & Trustworthiness & Informativeness & Satisfaction \\
    \midrule
    Grad-CAM & 5.99 ±1.14 & 5.91 ±1.03 & 5.62 ±1.13 & 5.02 ±1.71 & 5.57 ±1.43 \\
    LIME & 4.80 ±1.64 & 4.96 ±1.56 & 4.56 ±1.57 & 4.00 ±1.93 & 4.39 ±1.93 \\
    SHAP (Gradient Explainer) & 4.95 ±1.51 & 4.87 ±1.53 & 4.66 ±1.56 & 4.14 ±1.90 & 4.18 ±2.06 \\
    INV (Simplified) & \textbf{6.05} ±1.17 & 5.90 ±1.06 & \textbf{5.76} ±1.23 & 5.41 ±1.52 & 5.74 ±1.31 \\
    INV & 5.92 ±1.40 & \textbf{5.92} ±1.32 & 5.74 ±1.38 & \textbf{5.82} ±1.50 & \textbf{5.83} ±1.35 \\
    \end{tabular}
    \caption{The table presents the results of the comparative analysis between INVs and the state-of-the-art methods. Although our methods were slightly superior in nearly all aspects, statistical significance was only observed in \emph{informativeness}.}
    \label{tab:questionnaire_xai}
\end{table*}

\begin{figure}
    \centering
    \includegraphics[width=0.483\textwidth]{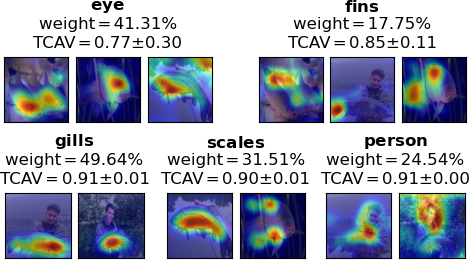}
    \caption{An example of a global explanation of the class ``tench'' for the last convolutional layer. It provides a set of features described by a label and associated with a weight and a TCAV score.}
    \label{fig:fish_global}
\end{figure}

\subsection{Towards Global Explanations}
Labeled cluster maps have the advantage of allowing the aggregation of local explanations into global ones. While our research primarily focuses on INVs as a local explainability technique, we propose a method for merging features from multiple images to produce class-level global explanations, intending to describe how the network generally recognizes a class. To create these global explanations, we aggregate the highest-scoring labels from all images within a class and a selected layer. Each label is assigned a weight as an indicator of feature importance, calculated as the average of the weights from the cluster maps for which it was the highest-scoring label. The visualization also includes cluster map examples for each feature whose label score was the highest. An example of a global explanation is shown in Figure \ref{fig:fish_global}, considering the class ``tench'' and the last convolutional layer.
To further validate the explanation, we also applied TCAV for the top five labels using 20-30 example images and 50 random runs. In the given example, we discovered that the TCAV score was high for all features identified by our method, with statistical significance (p-value $<$ 0.05). However, the ``eye'' feature scored slightly lower, despite our method attributing a high weight to it. This discrepancy could be because INVs compute the weights using the gradients of the pre-softmax score, while TCAV uses the gradients of the loss. It’s plausible that while the ``eye'' feature significantly contributes to the score for the class ``tench'', it does not reduce the loss as much because the feature is shared with other classes, such as the ``english springer''. On the other hand, the ``person'' feature is important for both methods, suggesting a slight overfitting in the ``tench'' class. Considering this, it may be beneficial to combine INVs and TCAV to achieve more robust global explanations as these methods can be viewed as complementary. Indeed, INVs can identify the features extracted by CNNs, while TCAV provides verification for their labels.

\subsection{A Discussion on Deep Reveal \label{questionnaire_res}}
The participants of the gamified activity also filled out a workload and usability questionnaire to validate \emph{Deep Reveal} and find possible improvements. The questions were taken from the System Usability Scale (SUS) \cite{Brooke}, and the NASA-TLX \cite{HART}. 
The final SUS score was 80.9/100, corresponding to a percentile ranking of 90, indicating that the system is considered between good and excellent \cite{Brooke_SUS}. The NASA-TLX score was 38.1/100. This score is considered somewhat high (the lower the score, the better) \cite{Prabaswari}, suggesting that \textit{Deep Reveal} still has room for improvement in workload aspects, more than in usability. One interpretation could be that guessing and labeling pictures requires effort since they are not easy tasks. Additionally, a significant amount of feedback underlined that some masked images were disproportionately more difficult to guess than others. This can be attributed to the fact that not all cluster maps focus on discriminative regions. A takeaway for future implementations is to design games with a more balanced difficulty, which could be achieved by revealing a slightly bigger portion of the image if the weight of the corresponding cluster map is low. Another approach could be introducing different difficulty levels based on the cluster map weight. This idea derives from users using slightly fewer hints when playing with more important cluster maps. Users also noted that it was sometimes challenging to assign labels to features they were unfamiliar with or didn't know the names of. This emphasizes the importance of including expert users with adequate knowledge about the classified entities, otherwise the explanations may become oversimplified. However, over-relying on experts may lead to an overestimation of the network's specificity.

%-----------------------------------------------------------------------------
% Conclusion
%-----------------------------------------------------------------------------
\section{Conclusion and Future Works \label{conclusions}}
In this article, we introduced Interpretable Network Visualizations, a human-in-the-loop approach for explaining image classification. Our results demonstrated the potential of our method while providing possible research directions and improvements. These improvements include extending \textit{Deep Reveal} to collect labels for images the network mispredicted, for example by considering only labels provided by users who made the same error as the machine. Future works could also explore the possibility of associating CNN filters with labels. This could enable the generation of an INV of an image simultaneously with the CNN execution, implying that the crowdsourcing step would only be required once. Moreover, future studies could investigate the integration of Large Language Models in our workflow. This would eliminate the need for crowdsourcing, resulting in fully automated, although less transparent, explanations.
Finally, it could be interesting to investigate the applicability of our method beyond CNN architectures and image classification.
\appendix

\section*{Ethical Statement}
The absence of transparency in Artificial Intelligence systems poses a serious ethical dilemma, particularly as Convolutional Neural Networks are being widely utilized in critical sectors such as healthcare and autonomous driving. Our research aims to contribute towards improving the transparency and interpretability of these models, thereby facilitating a more trustworthy adoption.

\section*{Contribution Statement}
Matteo Bianchi and Antonio De Santis contributed equally to this paper.

\bibliographystyle{named}
\bibliography{ijcai24}

\end{document}